\documentclass[preprints,article,accept,moreauthors,pdftex]{Definitions/mdpi} 
\usepackage{amsmath}
\usepackage{subfig}
\usepackage{booktabs}
\renewcommand{\arraystretch}{1.2}

\firstpage{1} 
\pubvolume{xx}
\issuenum{x}
\articlenumber{x}
\pubyear{2021}
\copyrightyear{2021}
\history{ }

\pdfoutput=1

\Title{Multi-Perspective Anomaly Detection}

\Author{Peter Jakob $^{1\ddagger}$\orcidA{}, Manav Madan $^{1\ddagger}$, Tobias Schmid-Schirling$^{1}$\orcidB{} and Abhinav Valada $^{2}$\orcidC{}}

\AuthorNames{Peter Jakob, Manav Madan, Tobias Schmid-Schirling, Abhinav Valada}

\address{%
$^{1}$ \quad Fraunhofer Institute for Physical Measurement Techniques IPM, Freiburg, Germany\\
$^{2}$ \quad Robot Learning Lab, University of Freiburg, Germany}

\firstnote{These authors contributed equally to this work.}

\abstract{Anomaly detection is a critical problem in the manufacturing industry. In many applications, images of objects to be analyzed are captured from multiple perspectives which can be exploited to improve the robustness of anomaly detection. In this work, we build upon the deep support vector data description algorithm and address multi-perspective anomaly detection using three different fusion techniques, i.e., early fusion, late fusion, and late fusion with multiple decoders. We employ different augmentation techniques with a denoising process to deal with scarce one-class data, which further improves the performance (ROC AUC $= 80\%$). Furthermore, we introduce the dices dataset, which consists of over 2000 grayscale images of falling dices from multiple perspectives, with 5\% of the images containing rare anomalies (e.g., drill holes, sawing, or scratches). We evaluate our approach on the new dices dataset using images from two different perspectives and also benchmark on the standard MNIST dataset. Extensive experiments demonstrate that our proposed {multi-perspective} approach exceeds the state-of-the-art {single-perspective anomaly detection on both the MNIST and dices datasets}. To the best of our knowledge, this is the first work that focuses on addressing multi-perspective anomaly detection in images by jointly using different perspectives together with one single objective function for anomaly detection.}


\begin{document}


\section{Introduction}

Abnormalities are present everywhere in day-to-day life. In many cases, these abnormalities or anomalies can be life critical and therefore they have to be identified at all costs. Anomaly detection is the process of identifying dissimilar samples from a total lot of samples~\cite{Zimek2017}. A wide range of applications already exists which are based on the detection of abnormalities such as medical diagnostics, video surveillance, fraud detection, surface defect detection, etc.~\cite{minhas2019anomaly}. It is fairly simple for humans to detect abnormalities in images if some prior information about what anomalies look like is known. However, this is an extremely complex task to automate. To begin with, the borderline between normal and anomalous behavior is often marginal and imprecise. The scarcity of labeled data for training and validation also imposes further limitations on the performance that can be achieved in anomaly detection. Two major challenges that are typically encountered in anomaly detection with high dimensional data such as images are the absence of labeled data and the
shortage of anomalous instances~\cite{minhas2019anomaly}. To address these problems, this work focuses on unsupervised anomaly detection algorithms that require no labels in the training set or only labels from one class. Algorithms that utilize only one class of samples come under the category of one-class learning algorithms that aim to identify whether a data point belongs to a particular category~\cite{khanMadden2014}.

To solve multi-perspective anomaly detection, one has to first know how single-view anomaly detection can be addressed. Standard single-view anomaly detection techniques such as One-Class SVM (OC-SVM)~\cite{scholkopf2000support} or Kernel Density Estimation (KDE)~\cite{parzen1962estimation} work out of the box with low-dimensional data but these techniques often fail when dealing with high-dimensional data such as images~\cite{Ruff2018}. To successfully use these methods with images, typically discriminative features or feature vectors of fixed length are extracted using some hand-designed feature extractors. This process is time-consuming and also requires human effort for designing sensible feature extractors.
Deep learning has established itself as the first and foremost method for automatic feature extraction from images. Single-view anomaly detection techniques can be broadly categorized into two types of methods, i.e., pure deep learning-based algorithms that solve anomaly detection, or a combination of deep learning with traditional machine learning approaches~\cite{chalapathy2018anomaly}. The latter approach is also known as a hybrid approach as it combines two domains by using the advantages of both. One such state-of-the-art pure deep-learning-based single-view anomaly detection technique is Deep Support Vector Data Description (Deep SVDD)~\cite{Ruff2018}. Deep SVDD can be considered as an extension of the Support Vector Data Description (SVDD)~\cite{tax2004support} algorithm which aims to confine all the one-class data points inside a minimal volume hypersphere. The Deep SVDD algorithm employs a neural network to transform high-dimensional input into lower-dimensional feature space such that a minimal volume hypersphere comprising of all the one-class data points can be built.

In this work, we extend the Deep SVDD algorithm to address multi-perspective anomaly detection based on novel fusion techniques. If multiple images of a sample are available, all having different pose or lighting conditions then how does one decide if the object is anomalous? For making such a prediction, one may need to know in which perspective the anomaly is present, and if it can be trusted. The multi-perspective anomaly detection task can be addressed using state-of-the-art single-perspective anomaly approaches by employing them multiple times on each of the perspectives, followed by a fusion of the individual predictions with a voting scheme. Here, a voting schema such as majority voting should consider the statistics of each experiment and their inter-dependency to obtain consensus. Motivated by the expected anomalies (e.g., scratches on a concave surface), an experimental setting with three perspectives can be designed where a majority schema will never be able to detect the anomaly as long as at most one perspective has the chance to view the anomaly. In contrast, a minority voting is not promising while considering finite false positive or negative rates, e.g., the pseudo scrap rate will be always higher compared to single-perspective runs. Consequently, we have to adapt an existing single-perspective approach and fuse 
information prior to classification.
To combine information from different perspectives, three different fusion strategies are developed, i.e., early fusion, late fusion, and late fusion with dual decoders.

We outline the major contributions of this work as:
\begin{itemize}
	\item We enhance the Deep SVDD algorithm by using a denoising process and data augmentation techniques.
	\item We introduce three different fusion approaches that can handle multiple perspectives (as a proof of concept we work with two perspectives) of an object to obtain one single robust prediction as to whether the object is anomalous.
	\item We develop a new multi-perspective image dataset (also containing two perspectives) that is used for evaluating different algorithms. 
	\item We perform exhaustive evaluations of our approach on both our proposed dices dataset {and the standard MNIST dataset} on which we achieve state-of-the-art performance. {For the dices dataset, the focus is on the detection of low-level anomalies. The use of MNIST refers to high-level anomalies, because the whole object (i.e., every single digit) changes. }
\end{itemize}

The rest of the paper is structured as follows. We first present the related work in anomaly detection in Section~\ref{sec_related}. We then present a short introduction on single-view anomaly detection technique Deep SVDD~\cite{Ruff2018} in Section~\ref{sec_mat}, followed by a detailed description of the three different fusion techniques. Subsequently, we describe the data collection procedure that we employ for the ``dices dataset'' that we introduce in this work and a brief description of how we extend the MNIST dataset~\cite{MNIST} to a multi-perspective setting. In Section~\ref{sec_res}, we present the results of our approach {on multi-perspective data} in comparison to baseline techniques and provide a thorough discussion on the results in Section~\ref{sec_dis}. Finally, we present the conclusions and directions for future work in Section~\ref{sec_con}.

\section{Related Work}
\label{sec_related}

Ruff~{et~al.}~\cite{Ruff2021} already covers an extensive review to detect anomalies with the help of machine learning and deep learning techniques. Two major findings are: First, deep learning can be easily applied to high-dimensional data. Second, one favorite route to extract features implies convolutional autoencoders of different flavors of which we are using deep (e.g., \cite{seeboeck2016}) and denoising autoencoders~\cite{kayser2015denoising}. Furthermore, generative approaches also try to learn the ``normal'' class~\cite{Goodfellow-et-al-2016} but are harder to train. In this contribution, we employ the Deep SVDD~\cite{Ruff2018} approach that uses the advantage of feature extraction with the help of an autoencoder. Furthermore, Ruff~{et~al.} even provide a semi-supervised version of their Deep SVDD~\cite{ruff2020deep}. If only one-class training data is available this variation of Deep SVDD cannot be employed. 

This contribution focuses on how to fuse information for multi-perspective anomaly detection. Multiple perspectives or views are generally associated with data obtained from different modalities~\cite{Ji2019}. In the case of high dimensional data such as images, it also refers to different features that can be extracted for a single-view such as a scene descriptor (GIST, \cite{Oliva2001}), Histogram of Oriented Gradients (HOG), and Local Binary Patterns (LBP)~\cite{Hou2020}. Whereas, multi-perspective in our case refer to different perspectives of an object that are captured from different orientations. The fusion of multiple perspectives for addressing complex tasks such as image classification or 3D shape reconstruction has gained interest in recent years~\cite{su2015multiview, choy20163dr2n2, LIN2018, zhu2019multi, Seeland2021}. The fusion of (low-dimensional) data is common in many fields such as autonomous vehicles where data from different sensors are merged using Kalman filters~\cite{subramanian2009sensor}, whereas the fusion of high dimensional data puts new questions.

First, how to fuse the information. Sophisticated approaches use so-called gating networks which compute averaging weights in inference times, e.g., \cite{kim2018robust, valada2016convoluted}. Another approach is to use recurrent networks to merge autoencoders~\cite{choy20163dr2n2}. As a robust training of anomaly detection itself is challenging, we simpler methods such as stacking images as different color channels~\cite{valada2016towards} can be exploited. {This is precisely one of the innovations of the present contribution. We achieve significant anomaly detection rates with simple fusion approaches.} Next, when do the data fusion~\cite{Seeland2021}., e.g., Sun~{et~al.} fuse their information almost at the beginning of one network~\cite{Sun2017} whereas Lin and Kumar~\cite{LIN2018} concatenate the compressed feature vectors at the end of individual processing. Seeland and Mäder~\cite{Seeland2021} pointed out that doing so (with the help of a fully connected layer) is the most promising in their classification tasks. {Thus, a novelty of this paper is not only to answer the question when the fusion will takes place but also to answer this question in an unsupervised setting.}

Nevertheless, there is a lack of work on anomaly detection using multiple views. One of the application covers the detection of facial micro-expressions~\cite{Sheng2019}, e.g., applied on the CASME dataset~\cite{Yan2013}. Sheng~{et~al.} extract features from images through local binary pattern (LBP) histograms and perform the anomaly detection step on them. {In contrast, Deep SVDD directly acts on the latent space. Therefore, it is desirable to build a multi-perspective variant of it to further improve its robustness.} 

\section{Materials and Methods}
\label{sec_mat}

There are two major traditional one-class unsupervised anomaly detection techniques namely One-Class SVM (OC-SVM)~\cite{scholkopf2000support} and SVDD~\cite{tax2004support}.
The OC-SVM aims to separate the majority of data points from the origin with a support vector classifier, i.e., a hyperplane.  On the other hand, SVDD uses a hypersphere as a classifier to include all one-class data points. As the basis of our multi-perspective approach is the Deep SVDD~\cite{Ruff2018} algorithm which in itself is an adaptation of the SVDD algorithm. {We briefly review the baseline methods (such as OC-SVM) and then describe the Deep SVDD approach.}

\subsection{Shallow One-Class Baseline}\label{sec_baseline}
{Non-deep (i.e., shallow) methods are suitable for comparing individual results of the present work against state-of-the-art methods. 
For this purpose, the image data are first reduced in their dimensions using Principal Component Analysis (PCA). Ruff et al. also proceed in this way and suggest that the total variability should not fall below 95\%. The following methods use these reduced data: OC-SVM \cite{scholkopf2000support}, Kernel Density Estimation (KDE), Isolation forest \cite{Liu2008} (IF). When executing and parameter configuring the methods, we also follow Ruff et al. \cite{Ruff2018}.}
{
Since OC-SVM is a kernel-based method, we choose the Gaussian radial basis function for it. Using grid search, we discretely search for optimal values for the inverse length scale $\gamma \in \left\{2^{-10},2^{-9},2^{-8},\dots,2^{-1}\right\}$ and $\nu \in \left\{0.01,0.05,0.1\right\}$. The latter controls the smoothness of the boundary between inliers and outliers.}
{Kernel Density Estimation is also performed using Gaussian kernel. The bandwidth is optimized using grid search within the elements $\left\{2^{0.5},2^{1.0},2^{1.5},\dots,2^{5}\right\}$. 
The Isolation Forest \cite{Liu2008} (IF) is an impressive example of how an anomaly can be understood as the path length of a decision tree. 100 base estimators are chosen in the ensemble.}

\subsection{Deep SVDD}

There are two major drawbacks of {Support Vector Data Description} \cite{tax2004support} (SVDD) when it comes to employing this approach on high-dimensional data such as images. First, due to the curse of dimensionality, SVDD with high dimensional data requires explicit feature engineering to perform dimensionality reduction. Secondly, SVDD is a kernel-based approach and hence requires an appropriate kernel function to perform feature transformation such that a minimal volume hypersphere can be built. Deep learning offers an appropriate solution for both of the above problems. Deep support vector data description (Deep SVDD) utilizes the advantages of the SVDD approach with deep learning and performs automatic feature extraction by employing a convolutional autoencoder (CAE) paradigm, compare Figure~\ref{fig_network}. This approach was introduced by Ruff~{et~al.}~\cite{Ruff2018} in 2018 and yields state-of-the-art results for single-view anomaly detection on public {datasets} such as MNIST~\cite{MNIST}.

\subsubsection{Training and Objective Functions}
\label{sec_svdd_training_and_objectives}

The original training of Deep SVDD is formulated as a two-stage approach, compare Figure~\ref{fig_fusion}. The first stage is a pre-training where discriminative features representing the one-class data are extracted. The objective function of this first stage is the pixel-wise reconstruction error comparing input and output images.

The second stage uses the learned weights of the encoder as weight initialization step and the decoder is discarded. These learned weights provide a jump-start for an encoder network. This network maps data from input space $\mathcal{X}\subseteq\mathbb{R}^d$ to feature space $\mathit{F}\subseteq\mathbb{R}^p$. The network is parametrized by a set of weights $\mathcal{W} = \{\mathit{W}^1,\dots,\mathit{W}^l\}$ where $\mathit{W}^l$ are the weights of the layer $\mathit{l} \in \{1,\dots,L\}$. The goal is to learn network weights $\mathcal{W}$, together with minimizing the volume of a data-enclosing hypersphere in feature space $\mathit{F}$. The hypersphere is characterized by radius $R > 0$  having a center $c \in \mathit{F}$ which lies in the same feature space $\mathit{F}$. The objective function is formulated as
\begin{eqnarray}\label{eq_ruff_obj}
	\operatorname*{min}_{\mathit{R},\mathcal{W}} \quad \mathit{R}^2 + \frac{1}{\nu\mathit{n}}\sum_{i=1}^n \mathit{max}\{0,\lVert \phi(x_i;\mathcal{W}) - c \rVert^2 - \mathit{R}^2\} + \frac{\lambda}{2} \sum_{l=1}^L\lVert \mathit{W}^l \rVert ^2_{{F}}.
\end{eqnarray}

The first term in the Equation~\eqref{eq_ruff_obj} refers to minimizing the radius $R$ of the hypersphere. The second term is a hinge loss~\cite{bishop2006pattern} that is used to obtain a soft boundary. Similar to SVDD, the hyperparameter $\nu \in [0,1]$ here also controls the trade-off between the volume of the sphere and violations of the boundary. These violations force the sphere to have a soft boundary that further results in robustness and better generalization. The last term is a weight decay regularizer on the network parameters $\mathcal{W}$.

Once the training is complete, the learned radius and center values are used to compute the anomaly score. For the soft boundary objective, this anomaly score for each  $x_i$ present in the test dataset is computed by evaluating
\begin{eqnarray}\label{eq_anomaly_score}
Anomaly Score = \lVert\phi(x_i;\mathcal{W}) - c\rVert^2 - \mathit{R}^2,
\end{eqnarray}
where $\phi(x_i;\mathcal{W})$ is the output of the second stage encoder network. $R$ and $c$ are the learned radius and center values obtained from the training set. {Input $x_i$ is deemed to be anomalous if the sign of $Anomaly Score$ is positive, i.e., $x_i$ lies outside the hypersphere. A negative sign indicates a non-anomalous input.}

\subsection{Network Architecture}
\label{sec_network}

\begin{figure}
	\centering
	\includegraphics[width=0.98\textwidth]{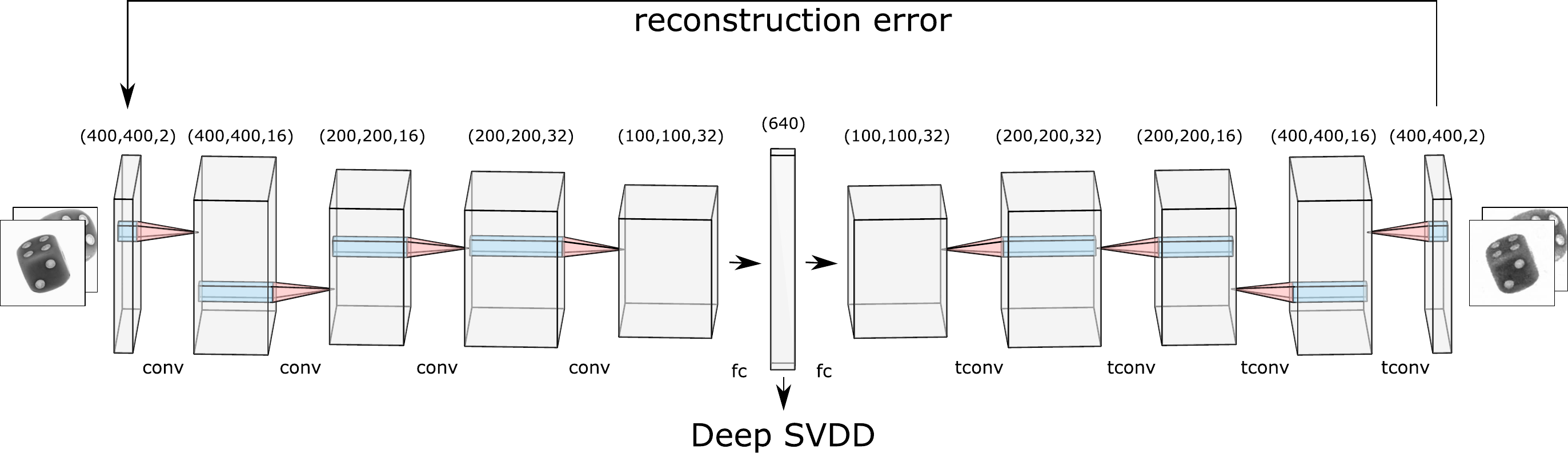}
	\caption{Illustration of the convolutional autoencoder used for multi-perspective anomaly detection. There are five layers each in the encoder and decoder block. Each layer can be identified with a particular name (``conv'' for convolution, ``tconv'' for transpose convolution, and ``fc'' for fully connected) and the size that it outputs. The arrow towards Deep SVDD portrays that this is used as the feature space where the hypersphere is found. Figure partly generated with LeNail's schematics \cite{LeNail2019}.}
	\label{fig_network}
\end{figure}

Figure~\ref{fig_network} illustrates how two stacked 2D images of dices are fed as input to the network. The output dimension of each layer is depicted on the top of each block. Initially, a cubic feature map is extracted from the input image. This feature map then becomes the input for the next layer that outputs a cubic-shaped feature map and so forth. In the middle, there are two fully connected layers where the first one maps the flattened output of the last convolution layer to a dimension of 640. This number represents the size of the latent space and is a hyperparameter that needs to be optimized. The latent space holds all the encoded features in a lower-dimensional space. The next fully connected layer maps this input latent space back to the dimension of the output of the last convolution layer. Then, this 1D output generated from the fully connected layer is reshaped into a 3D tensor so that the transposed convolution operation can be employed. This decoding process converts the features in the latent space back to the stacked 2D images using the transposed convolutional layers.


\subsection{Deep SVDD with Fusion Techniques}
\label{sec_fusion}

In this work, we extend the single-view Deep SVDD approach for multi-perspective anomaly detection. One major requirement for this is to process multiple perspectives for a single object simultaneously. We address this problem by using three different fusion techniques that are based on the fusion of information either at the input state or in the latter part, i.e., in the feature space $\mathit{F}$. For proof of concept, we use two different perspectives for the formulation of (the) fusion techniques but these can be readily extended to more than two perspectives.


\begin{figure}
\includegraphics[width=0.97\textwidth]{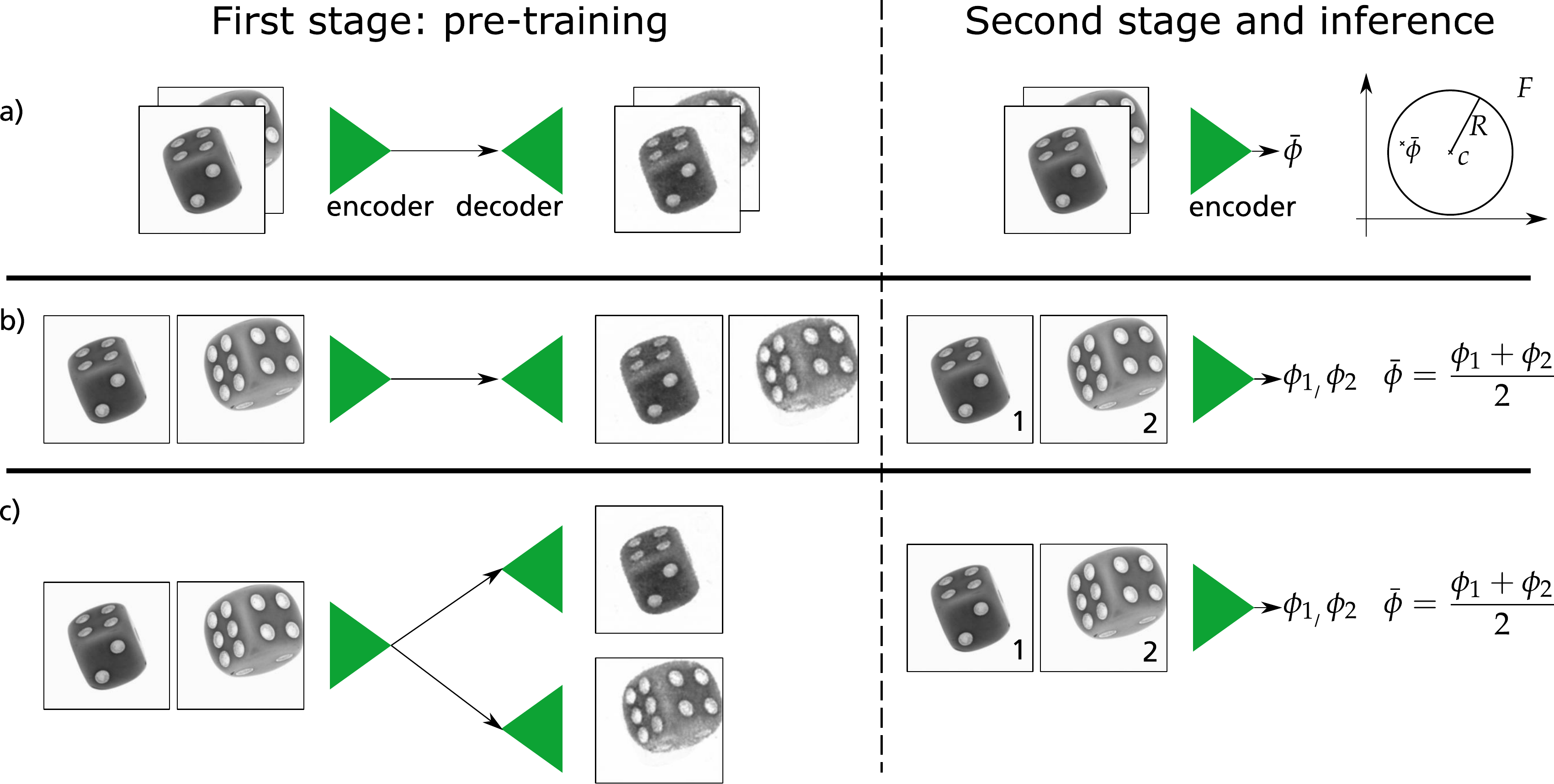}
\caption{Given two perspectives of the same dice, this figure illustrates the training and fusion techniques that we use in this work. Left column: The first stage comprises of the auto encoder training. Right column: The weights of the encoder are then further trained alone according to Equation~\eqref{eq_ruff_obj}. (\textbf{a}) With an \textit{early fusion} the two perspectives were stacked channel-wise. Hence the output of the encoder comprises information of both perspectives inherently. (\textbf{b},\textbf{c}) In contrast to that, two outputs were averaged in a \textit{late fusion}. (\textbf{c}) The last row shows the utilization of two decoders. }\label{fig_fusion}	
\end{figure}

\subsubsection{Early Fusion}

The goal of this technique is to fuse information at the beginning of the algorithm, i.e., at the input stage. For this reason, different perspectives are stacked together channel-wise forming a multi-channel image. {Figure~\ref{fig_network} shows this process at the input of the encoder network. As described in Section~\ref{sec_svdd_training_and_objectives}, the goal of the first training stage is to reconstruct this stacked image pixel by pixel.} Hence, the CAE can handle multi-channel input and extract common features together from both perspectives. In the second training stage stage {(cf. Figure~\ref{fig_fusion})}, the stacked image is passed through {the} encoder network {only} which results in a single vector $\bar{\phi}$ representing a single point for both of the perspectives in feature space $\mathit{F}$. {Analogous to Ruff~et~al.~\cite{Ruff2018} the weights of this encoder are now updated until all vectors $\bar{\phi}$ of the respective stacked images in the training dataset are in a hypersphere, are as small as possible. This learned hypersphere is described with the parameters center $c$ and radius $R$. Finally, a vector $\bar{\phi}$ for a stacked image of the test dataset can be calculated with these trained parameters in inference time. Together with Equation~(\ref{eq_anomaly_score}), the $Anomaly Score$ is obtained, i.e., the statement whether the particular data point is inside or outside the learned hypersphere. Elegantly, this score is valid for both images in the stack at the same time - a final averaging of two anomaly scores is thus obsolete because exact one fused anomaly score is given.}

\subsubsection{Late Fusion and Late Fusion with Dual Decoders}

For the late fusion, we propose two different techniques, i.e., late fusion {w/ and w/o} dual decoders. As the name suggests, in both of these fusion techniques the information is fused in the latter part of the algorithm. Similar to the early fusion {from the preceding section} , the late fusion techniques also have two stages of training ({cf.} Figure~\ref{fig_fusion}). 
{In the first training step for late fusion w/o dual decoders, a CAE is trained to reproduce the input image as well as possible regardless of the perspective from which the images were captured. In contrast, two CAEs are trained for the approach late fusion w/ dual decoders where the weights of the encoders are shared. Depending on the perspective, either one or the other CAE is trained.}
The goal is to have one feature extractor network for all the different perspectives. {In contrast to early fusion from the preceding section}, different perspectives are not stacked together.
In the second training stage, as shown in Figure~\ref{fig_fusion}, each {image of the training dataset} is forwarded one at a time through the encoder network. {The aim is to again map all feature vectors as close together as possible. For this purpose, the weights of the encoder are iteratively updated according to the cost function in \mbox{Equation~(\ref{eq_ruff_obj})}. The result of this optimization is a hypersphere which is as small as possible with the parameters radius $R$ and center $c$. At \textit{inference time}, a data point $x_i$ of the test dataset can be pushed through the encoder to calculate a corresponding feature vector $\phi(x_i;\mathcal{W})$. Since this feature vector belongs to a single perspective, the feature vectors of two perspectives have to be arithmetically averaged according to }
\begin{eqnarray}\label{eq_late_fusion_average}
\bar{\phi} = \frac{1}{2}\left(\phi(x_1;\mathcal{W})+\phi(x_2;\mathcal{W})\right).
\end{eqnarray}

Finally, the anomaly score is then computed using Equation~\eqref{eq_anomaly_score} where $\phi(x_i;\mathcal{W})$ is the averaged feature vector {$\bar{\phi}$} obtained from {Equation~(\ref{eq_late_fusion_average})}. 

\subsection{Hyperparameter Optimization}
\label{sec_hyp_opti}

There are different hyperparameters that one has to optimize for the two different stages of multi-perspective Deep SVDD. The gradient descent algorithm is used to optimize the weights $\mathcal{W}$ of the neural networks by utilizing the backpropagation algorithm~\cite{bishop2006pattern}. 
For the {\textit{pre-training stage}}, one has to find an optimal architecture for the autoencoder and also select an optimal loss function as well as an optimizer. The hyperparameters associated with the architecture of any CAE include the number of convolutional and deconvolutional layers in the encoder and decoder respectively. This is further followed up by setting the kernel sizes of each layer. The encoder downsamples the incoming high-dimensional input to a reduced lower-dimensional space know as the feature space. This feature space is also another important hyperparameter of CAE. For Deep SVDD, this feature space also defines the dimension of space in which the hypersphere is built.
Choosing the right values for the hyperparameters for any algorithm is a challenge because one has to determine which is the most significant metric that those hyperparameters should optimize. {To ensure comparability with other studies (e.g., \cite{Ruff2018,chalapathy2018anomaly}), we use ROC AUC as the cost function. We extend this cost function by the f1 score while handling an imbalanced test dataset of MNIST in Section~\ref{sec_res_mnist}}. The encoder network in the {\textit{second training stage}} is kept the same as the earlier found optimal encoder. The trained weights of the encoder are used as initialization for this network. The output of this network is a vector $\phi(x_i;\mathcal{W})$ in feature space. As network weights $\mathcal{W}$ and radius $R$ lie on different scales~\cite{Ruff2018}, their optimizing can be done in a warm-up manner, i.e., fixing $R$ for the first $k\in{}N$ epochs where $N$ is the total training iterations and updating $\mathcal{W}$ on each epoch and then updating $R$ with $\mathcal{W}$ on every epoch once the few $k$ epochs are over. The parameter $\nu$ (see Equation~\eqref{eq_anomaly_score}) which is used to control the trade-off between the outliers and the number of samples that support the boundary (support vectors) is optimized for each dataset. For hyperparameter optimization, we use the Bayesian optimization (BO) and Hyperband (HB) (BOHB)~\cite{falkner2018bohb} library which utilizes past experiences to make better decisions. While optimizing the total number of iterations, the minimum and maximum budgets are kept unchanged for all the different techniques to enhance comparability between results. These budget parameters decide how many iterations are run before the best set of hyperparameter values are found.
During the experiments, one can observe different results for different runs of the algorithm and to tackle these variations, the random number generator, i.e., seed value is fixed for the optimization. However, once the optimization is complete, the algorithms are run on different seed values to obtain the average performance in terms of the ROC AUC score.
    
\subsection{Evaluation Datasets}\label{sec_datasets}

To evaluate the one-class learning algorithm, a test dataset that contains both the classes, i.e.,~anomalous and non-anomalous samples is required. Generally, the non-anomalous samples are likely available but anomalous samples are hard to acquire. {In the literature (e.g., \cite{Ruff2018}), datasets for other classification tasks are used in the form that only individual classes are considered non-anomalous and the others anomalous, as we do later in this work with MNIST~\cite{MNIST}. This approach neglects the fact that the multi-perspective images of an object contain some redundancy. Hence, we create a multi-perspective dices dataset to train and evaluate our fusion techniques.}

\subsubsection{Dices Dataset}
\label{sec_dices_dataset}

\begin{figure}
	\centering
	\subfloat[ ]{\includegraphics[width=0.35\textwidth]{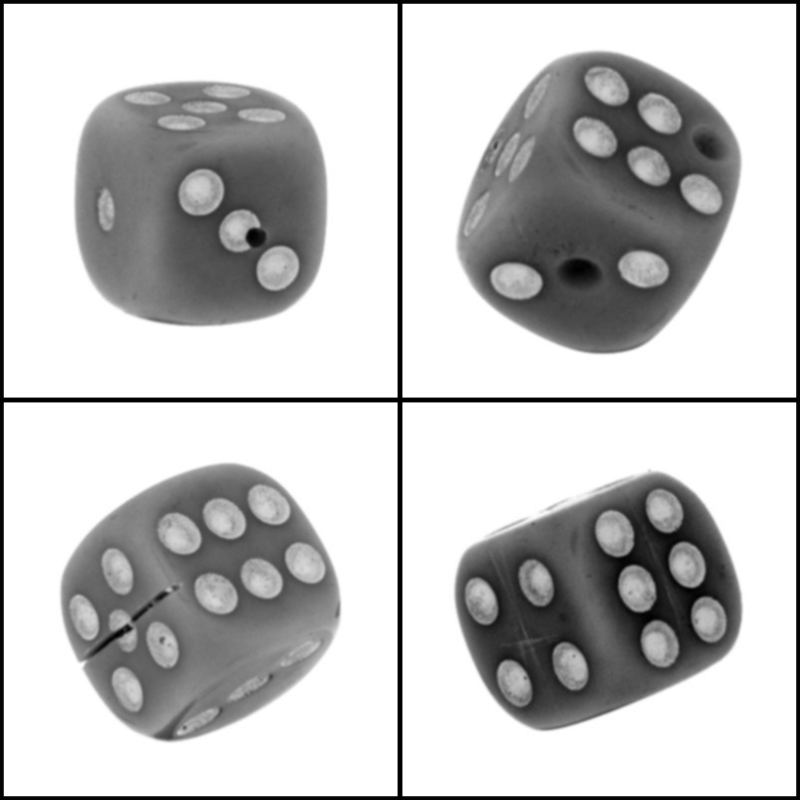}}
	\hspace{6pt}
	\subfloat[ ]{\includegraphics[width=0.35\textwidth]{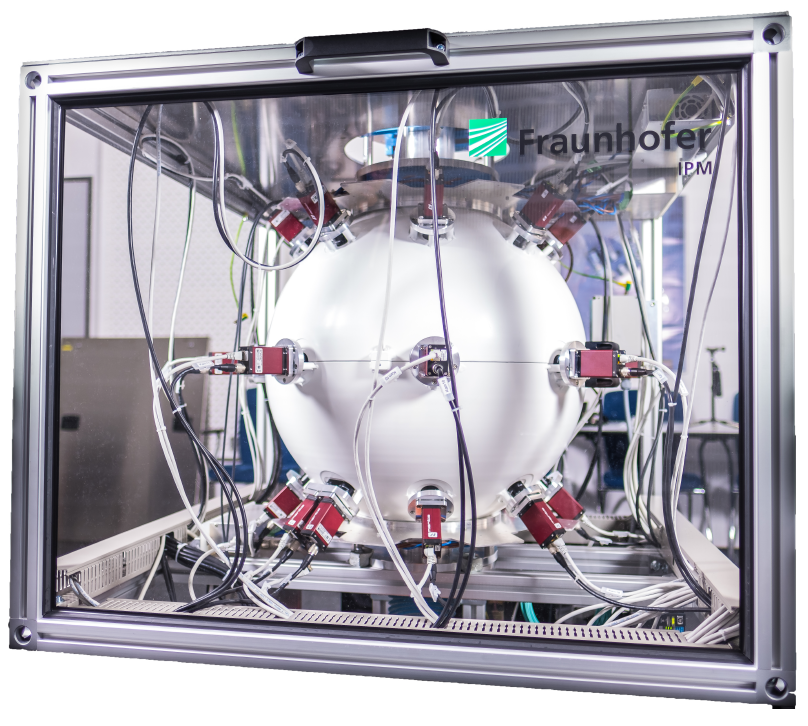}}
	\caption{(\textbf{a}) Four different types of anomalies were generated for dices: Drilling {(17\% occurances in test dataset)}, missing dots {(33\%)}, sawing {(17\%)} and scratching {(33\%)}. (\textbf{b}) The measurement setup free-fall~\cite{Jakob2019} where all 27 cameras are directed to the center of the globe.}
	\label{fig_free_fall}%
\end{figure}

The Free-Fall~\cite{Jakob2019} setup (see Figure~\ref{fig_free_fall}b) at Fraunhofer Institute for Physical Measurement Techniques (IPM) is used for data acquisition. The measurement setup has up to 27 monochromatic cameras taking images of the sample simultaneously with 20~\textmu{}s of illumination time. All 27 cameras are located at fixed relative positions from each other. The object is thrown from the top and all 27 cameras capture different perspectives simultaneously. For the multi-perspective dataset, the two most opposite perspectives are selected in terms of their pose. 
Choosing an object for creating a new dataset is a challenging task. For solving this, a few requirements are set which the object must fulfill such as the chosen object should be, e.g., readily available, and inexpensive.


Dices fit all these requirements perfectly. A traditional dice of a cubic structure is selected which has circles representing numbers from one to six on its faces.  A total of 11 dices of an edge size $16$~mm are selected out of which five randomly chosen dices are used for creating anomalies and six are set apart as non-anomalous samples. In \mbox{Figure~\ref{fig_free_fall}a}, one out of the total 27 perspectives for each anomalous dice is illustrated. All 27~images captured using a free-fall setup are high-resolution single-channel images having a dimension of $2464 \times 2056$ pixels each. Dice is cropped out (reducing the resolution to $400 \times 400$ pixels) of the image using the OpenCV library’s blob detection function. A total of 2000 (all from non-anomalous class) experiments (1 experiment = 27 perspectives) for the training dataset and 200 (100 non-anomalous and 100 anomalous) for the test dataset are performed by throwing respective dices through the apparatus. One has to note that there are only 11~physical available dices. 

For the multi-perspective training dataset, the two most opposite perspectives for all the 2000 non-anomalous experiments are selected and stacked channel-wise. For the test dataset, images from the very same perspectives are chosen. As the pose of the falling dice is not constrained in each experiment, these two particular perspectives have the chance to see anomalies at all in 60 out of 100 experiments. Hence the test dataset contains 60 anomalous samples. This fact emphasizes that more than two perspectives should be considered in the future for robust anomaly detection in this application. The dataset is openly available in \cite{MadanDices}. Table~\ref{tab_dataset} provides an overview of the dataset.


\begin{table}
	\centering
	\caption{Overview of our dices dataset~\cite{MadanDices}.}
	\label{tab_dataset}
	\setlength{\tabcolsep}{3mm}

	\renewcommand{\arraystretch}{1.1}
	\begin{tabular}{m{3.2cm} m{4cm}  m{4cm}m{1.6cm}}
		\toprule
		\textbf{Dataset} & \textbf{No. of Training Samples} & \textbf{No. of Test Samples} & \textbf{Resolution} \\
		\midrule
		Multi-perspective dices   &  2000, only 1 class, channel wise stacked non-anomalous samples & 73 non-anomalous + 60 anomalous channel wise stacked samples &  (400, 400, 2) 
 \\
		\bottomrule
	\end{tabular}
\end{table}

\begin{figure}
	\centering
	\subfloat[ ]{\includegraphics[width=0.2\textwidth]{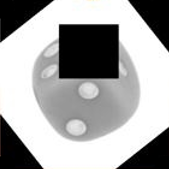}}
	\hspace{8pt}
	\subfloat[ ]{\includegraphics[width=0.2\textwidth]{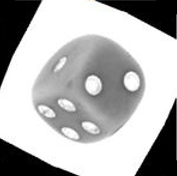}}
	\hspace{8pt}
	\subfloat[ ]{\includegraphics[width=0.2\textwidth]{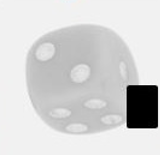}}
	\hspace{8pt}
	\subfloat[ ]{\includegraphics[width=0.2\textwidth]{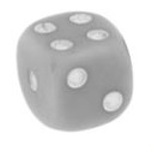}}
	
	\caption{Four additional training sets are created that extend the original data with augmentation. Each set consists of 4000~channel wise stacked images (400,400,2). (\textbf{a}) The first set contains all augmentation strategies, i.e., erasing patches, changing image constituents like brightness and altering geometry like flipping. The other training sets exclude particular strategies: (\textbf{b}) w/o erased patching, (\textbf{c}) w/o change of image constitution and (\textbf{d}) w/o geometrical rotations and flipping.}
	\label{fig_augm}%
\end{figure}

Certain applications in the industry often have difficulties in acquiring significant amounts of data. To address this problem, four different sets of augmented multi-perspective datasets are created additionally. These are summarized in Figure~\ref{fig_augm}. In these training datasets, the original stacked images are included together with 2000 additional augmented stacked images. All the augmentations are generated randomly for each image based on the set. 

Augmentations are characterized in three different categories:
\begin{itemize}
	\item[$\bullet$]Hiding information: Randomly erasing a patch (random area) from that image.
	\item[$\bullet$]Changing image constituents: Randomly changing the brightness, contrast, and saturation of an image or adding Gaussian noise.
	\item[$\bullet$]Altering geometry: Randomly flipping the image vertically or horizontally. Rotating the image at some random degrees.
\end{itemize}

The first training dataset contains all the above-mentioned transformations which are randomly employed on each image. In the second set, augmentations other than erasing patches are employed. In the third set, augmentations other than geometrical effects such as randomly flipping the image vertically or horizontally are employed. In the last set, augmentations other than changes in image constituents such as saturation, brightness, or contrast are used.

Furthermore, to improve the reconstruction quality of CAE, one can use denoising autoencoders (DAEs))~\cite{kayser2015denoising} where the input is corrupted by adding noise. This noisy input is then given to the autoencoder to obtain a reconstructed image but the loss is computed between the original (without noise) input and the reconstructed image. Such denoising process forces the network to extract discriminative features irrespective of the fact that there is random noise in the input which it will have to remove to have good reconstruction.

\subsubsection{MNIST Dataset}

The MNIST~\cite{MNIST} dataset of handwritten digits can be used for anomaly detection by considering one digit at a time as non-anomalous and the other as anomalous. The resolution of the images in the dataset is kept similar to the dices dataset, i.e., \mbox{$28 \times 28$ pixels} are upscaled to $400 \times 400$ pixels. A total of 4000 images per digit are selected as training samples for the non-anomalous class which is stacked in two channels to form the training set of size (2000, 400, 400, 2) per digit. For the test dataset, 400 samples are selected in total, i.e., 40 samples from non-anomalous digits and the rest from other digits. {For the anomalous case, different digits (other than the normal class) are randomly selected to make one single data point.} This gives unbalanced test sets which will be considered in the optimization. In total, there are ten different {datasets} where one particular digit is seen as non-anomalous. 

\section{Experimental Results}
\label{sec_res}

This section presents the results for single-perspective and multi-perspective anomaly detection. For running these different models, a local workstation is used which is equipped with NVIDIA Quadro RTX 6000 GPU. Metric functions such as ROC AUC score, Precision and Recall~\cite{Goodfellow-et-al-2016} are used for evaluating the performance of each algorithm (see \mbox{Appendix~\ref{sec_calc_metrics}}). The PyTorch framework is used for building all these different models.

\subsection{Multi-Perspective Anomaly Detection}\label{sec_res_multi_persp}

As discussed in Section~\ref{sec_svdd_training_and_objectives}, the training phase of Deep SVDD involves the training of two neural networks sequentially. The hyperparameters from both of the stages are optimized for the performance observed for anomaly detection. The autoencoder comprises of an encoder (four layers of convolutional layers and a fully connected layer) and a decoder (four layers of transposed convolutional layers and a fully connected layer) whereas the number of convolutional filters per layer and size of latent space is optimized individually for each fusion technique. The weights of the autoencoder are initialized using Xavier initialization. No bias is used in any of the two networks as the networks with bias terms can easily learn any constant function. {Ruff et al. \cite{Ruff2018} show that the problem in terms of a so-called hypersphere collapses when bias terms lead to a constant center of the sphere independent from the input.} The hyperparameters for the best configuration found using BOHB are listed in Appendix~\ref{sec_app_hyper_multi}. The results obtained with fusion techniques on our multi-perspective dices dataset are presented in Table~\ref{tab_res_multi_persp}.

\begin{table}
	\centering
	\caption{Results for different fusion techniques on our multi-perspective dices dataset. The ROC AUC values are averaged over three experiments which were run with different seed values. True positive samples (TP) are anomalous, etc.}
	\label{tab_res_multi_persp}
	\renewcommand{\arraystretch}{1.2}
	\begin{tabular}{m{3cm}m{2.5cm}m{1.6cm}m{1.6cm}m{3.03cm}}
		\toprule
		\textbf{Techniques} & \textbf{ROC AUC} & \textbf{Precision} & \textbf{Recall} & \textbf{Confusion Matrix}\\
		& \textbf{(Averaged)} & & & \Big($\begin{smallmatrix}
			     \mbox{\textbf{TN}} & \mbox{\textbf{FP}} \\
			     \mbox{\textbf{FN}} & \mbox{\textbf{TP}}
\end{smallmatrix}$\Big)\textbf{ (\%)}\\
		\midrule
		Early fusion& 0.630 & 0.607 & 0.604 &
		$\bigl(\begin{smallmatrix}
69&31\\48&52
\end{smallmatrix} \bigr)$\\	
		Late fusion& 0.583 & 0.659 & 0.625 &
		$\bigl(\begin{smallmatrix}
84&16\\60&40
\end{smallmatrix} \bigr)$\\
		\mbox{Late~fusion w/} \mbox{dual decoders} &0.561&0.226&0.5&
		$\bigl(\begin{smallmatrix}
0&100\\ 0&100
\end{smallmatrix} \bigr)$\\
		\bottomrule
	\end{tabular}
\end{table}

Forcing the network to remove random noise while learning discriminative features is critical for anomaly detection. This factor is visible in Table~\ref{tab_res_multi_persp_denoise} which shows the improvement in multi-perspective anomaly detection results when a denoising autoencoder is employed instead of the standard autoencoder in the pre-training stage. We observe that the early fusion technique is performing the best in terms of the ROC AUC score when compared to other fusion techniques. The worse performance in terms of precision and recall is achieved with a late fusion w/ dual decoders. One has to bear in mind that the optimization of hyperparameters takes ROC AUC scores as a cost function and not, e.g., average precision. {Moreover, we also compare with the single-perspective Deep SVDD that only uses one perspective, without the fusion of multiple perspectives. We use the same hyperparameters as the late fusion approach (see Table \ref{tab_app_hyperparameter_multi_view}, late fusion) as both the single-perspective model and one of the streams of the late fusion model have the same configuration, taking one perspective as input. The single-perspective model achieves} {a ROC AUC of 61.4\% (cf. Table~\ref{tab_res_multi_persp_denoise}) out of three runs.} {This result demonstrates the utility of multi-perspective anomaly detection, as the multi-perspective late fusion approach achieves an improvement.}

\begin{table}
	\centering
	\caption{Results for different fusion techniques on our multi-perspective dices dataset with a denoising autoencoder in the pre-training stage of Deep SVDD~\cite{Ruff2018}. True positive samples (TP) are anomalous, etc. {For the Late fusion technique, it is convenient to apply Deep SVDD to single images of the dices dataset---without taking into account the different perspectives. The result for this ``single-perspective'' Deep SVDD is given} {in the first row.}}
	\label{tab_res_multi_persp_denoise}
	\renewcommand{\arraystretch}{1.2}
	\begin{tabular}{m{3cm}m{2.5cm}m{1.6cm}m{1.6cm}m{3.03cm}}
		\toprule
		\textbf{Techniques} & \textbf{ROC AUC} & \textbf{Precision} & \textbf{Recall} & \textbf{Confusion Matrix}\\
		& & & & \boldmath{$\Big(\begin{smallmatrix}
			     \mbox{\textbf{TN}} & \mbox{\textbf{FP}} \\
			     \mbox{\textbf{FN}} & \mbox{\textbf{TP}}
\end{smallmatrix} \Big)$} \textbf{(\%)}\\
		\midrule
		{Single perspective} & {0.614} & {0.612} & {0.585} &
		{$\bigl(\begin{smallmatrix}
83&17\\ 66&34
\end{smallmatrix} \bigr)$}\\
    Early fusion& 0.740 & 0.638 & 0.621 &
		$\bigl(\begin{smallmatrix}
43&57\\ 19&81
\end{smallmatrix} \bigr)$\\	
		Late fusion & 0.697 & 0.659 & 0.660 &
		$\bigl(\begin{smallmatrix}
61&39\\ 29&71
\end{smallmatrix} \bigr)$\\
		\mbox{Late fusion w/} \mbox{dual decoders} &0.591&0.225&0.500&
		$\bigl(\begin{smallmatrix}
0&100\\ 0&100
\end{smallmatrix} \bigr)$\\
		\bottomrule
	\end{tabular}
\end{table}

{Furthermore, the results of the OC-SVM, KDE and IF methods (see Section~\ref{sec_baseline}) are presented as shallow baselines. For this purpose, the multi-perspective dices data are reduced with respect to dimensions by means of PCA and then transferred to the respective algorithms. After the grid search, the results in Table~\ref{tab_res_multi_baseline} are obtained. For the given parameters and models, only Deep SVDD w/ early fusion and KDE were suitable to obtain reasonable ROC AUC values.}

	\begin{table}
		\centering
		\caption{{ROC AUC results for anomaly detection on multi-perspective dices dataset using the denoising early fusion technique (Deep SVDD w/ EF). In addition, the results of the shallow anomaly detection methods OC-SVM, KDE and IF (see Section~\ref{sec_baseline}). The performance of Deep SVDD and IF are average values for 3 and 60 experiments, respectively.}}
		\label{tab_res_multi_baseline}
		\renewcommand{\arraystretch}{1}
		\begin{tabular}{m{3.5cm}m{3.15cm}m{3cm}m{2.5cm}}
			\toprule
		 {\textbf{Deep SVDD w/ EF}} & {\textbf{OC-SVM}} & {\textbf{KDE}}& {\textbf{IF}}\\
			\midrule
			
			 {\textbf{0.740 
}} & {$\le$0.500} & {0.673} & {$<$0.500}\\
			\bottomrule
		\end{tabular}
	\end{table}

\subsection{Multi-Perspective Anomaly Detection with MNIST}
\label{sec_res_mnist}

This section presents the results achieved on the standard MNIST~\cite{MNIST} dataset which we extended to the multi-perspective setting to evaluate the performance of the best fusion technique from the previous section, i.e., early fusion. All different ten datasets (one dataset per digit) are run using the denoising early fusion technique from the previous section. The hyperparameters regarding the training of the technique are optimized individually for each dataset using BOHB~\cite{falkner2018bohb}. The tuning parameters for BOHB are kept the same for each digit to compare the results with each other. For these experiments, the best configuration is not only selected based on the maximized ROC AUC score but also based on the {f1 score}. This is done due to the presence of a highly unbalanced test dataset (10\% images are from the non-anomalous class and the others are from the anomalous class) where the ROC AUC score could be misleading. 

\begin{figure}
		\begin{minipage}{.5\linewidth}
			\centering
			\subfloat[]{\label{a:}\includegraphics[scale=.3]{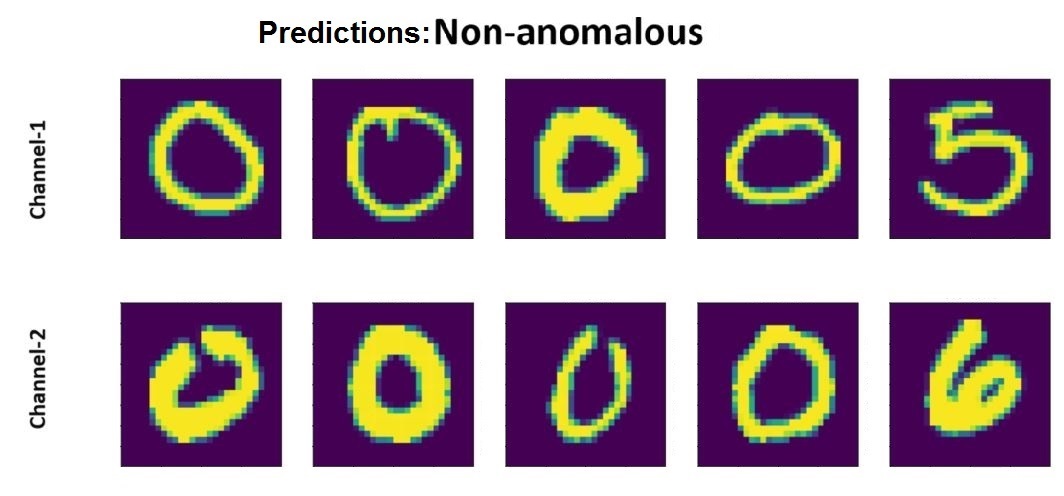}}
		\end{minipage}
		\hspace{-30pt}%
		\begin{minipage}{.5\linewidth}
			\centering
			\subfloat[]{\label{b:}\includegraphics[scale=.36]{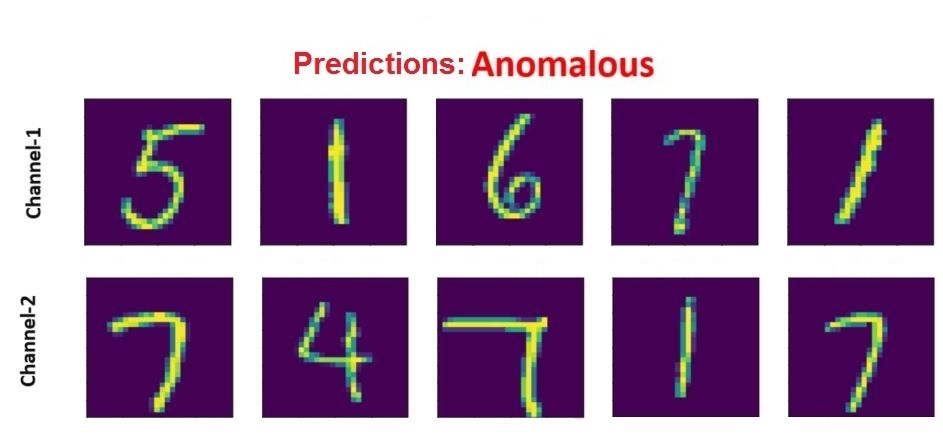}}
		\end{minipage}\par\medskip
		\centering
		\subfloat[]{\label{c:}\includegraphics[scale=.7]{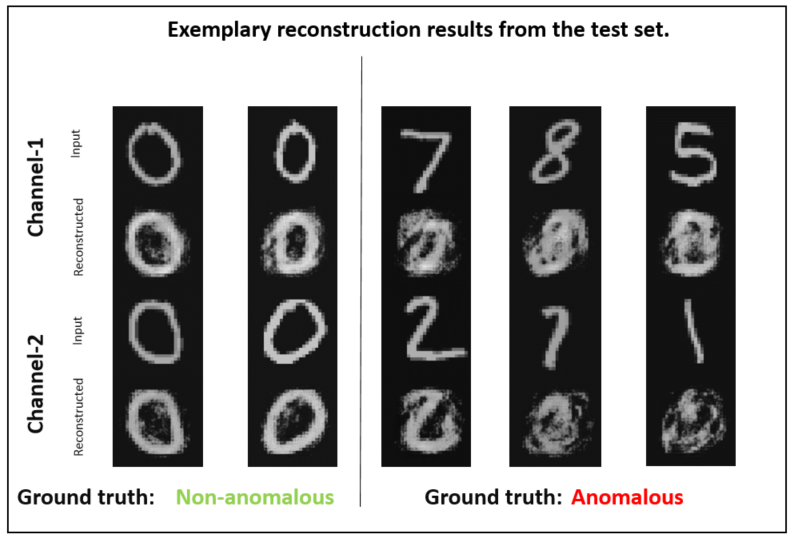}}
		\caption{Most normal (\textbf{a}) and most anomalous (\textbf{b}) samples determined by multi-perspective denoising fusion technique for multi-perspective MNIST~\cite{MNIST} dataset. In this dataset, samples representing zero are considered to be from non-anomalous class and other digits are considered to be from the anomalous class. {(\textbf{c})} illustrates the reconstructed output for examples from the test set.}
		\label{fig_mnist_inference}
	\end{figure}

The results achieved are presented in Table~\ref{tab_res_mnist}. Good performance is observed for all the multi-view MNIST dataset with denoising early fusion technique. In Figure~\ref{fig_mnist_inference}, the samples judged as anomalous and non-anomalous are illustrated for the dataset that has images representing digit zero as a non-anomalous sample. From Figure~\ref{fig_mnist_inference}a,b it can be analyzed that the denoising early fusion technique helps in predicting the anomalous samples and non-anomalous samples. The images in Figure~\ref{fig_mnist_inference}c represent some samples from the test dataset and it is visible that after the training, the non-anomalous samples are easily reconstructed by the autoencoder. Furthermore, the images of zeros are reconstructed easily with great detail. Whereas, for the other digits, the autoencoder fails to yield \mbox{significant results.}

{With the numbers in Table \ref{tab_res_mnist} we show an predominant independency of our approaches to almost all objects (digits) of the MNIST dataset as in the majority of cases there is a significant improvement in the anomaly detection rate. }

{To fit the figures shown into existing state-of-the-art methods, we pick out the cases of digits 2 and 3, as these seem to represent difficult cases with lowest ROC AUCs. The stacked MNIST data are now fed to a PCA rather than an autoencoder (see Section~\ref{sec_baseline}). The subsequent analysis using OC-SVM, KDE and IF lead to the results in Table \ref{tab_res_mnist_baseline}. }
	\begin{table}
		\centering
		\caption{Results for anomaly detection on multi-perspective MNIST~\cite{MNIST} dataset using the denoising early fusion technique. True positive samples (TP) are anomalous, etc. Additionally, the relative changes of the ROC AUCs w.r.t. to the original Deep SVDD \cite{Ruff2018} are shown.}
		\label{tab_res_mnist}
		\setlength\extrarowheight{1pt}
		\renewcommand{\arraystretch}{1.2}
		\begin{tabular}{m{4cm}m{2.4cm}m{1.15cm}m{1.18cm}m{3cm}}
			\toprule
			{\textbf{Non-Anomalous Digits}
} & \textbf{ROC AUC} & \textbf{Precision} & \textbf{Recall}& \textbf{Confusion Matrix}\\
			& \textbf{(Change to \cite{Ruff2018})} & & & \boldmath{$\Big(\begin{smallmatrix}
			     \mbox{\textbf{TN}} & \mbox{\textbf{FP}} \\
			     \mbox{\textbf{FN}} & \mbox{\textbf{TP}}
\end{smallmatrix} \Big)$} \textbf{(\%)}\\
			\midrule
			0 & 0.961 ($-$1.7\%) & 0.812 & 0.853 &$\bigl(\begin{smallmatrix}
75&25\\ 4&96
\end{smallmatrix} \bigr)$\\
			1 & >0.999 (+0.3\%)& 0.763 & 0.950 &
			$\bigl(\begin{smallmatrix}
100&0\\ 10&90
\end{smallmatrix} \bigr)$\\
			2 & 0.948 (+5.9\%)& 0.801 & 0.689 &
			$\bigl(\begin{smallmatrix}
40&60\\ 2&98
\end{smallmatrix} \bigr)$\\
			3 & 0.941 (+4.2\%)& 0.899 & 0.769 &
			$\bigl(\begin{smallmatrix}
55&45\\ 1&99
\end{smallmatrix} \bigr)$\\
			4 & 0.966 (+3.0\%)& 0.737 & 0.897 &
			$\bigl(\begin{smallmatrix}
90&10\\ 11&89
\end{smallmatrix} \bigr)$\\
			5 & 0.956 (+11\%)& 0.693 & 0.856&
			$\bigl(\begin{smallmatrix}
85&15\\ 13&87
\end{smallmatrix} \bigr)$\\
			6 & 0.973 ($-$0.7\%)& 0.886 & 0.936 &
			$\bigl(\begin{smallmatrix}
90&10\\ 2&98
\end{smallmatrix} \bigr)$\\
			7 & 0.979 (+5.6\%)& 0.878 & 0.791 &
			$\bigl(\begin{smallmatrix}
60&40\\ 1&99
\end{smallmatrix} \bigr)$\\
			8 & 0.954 (+2.7\%)& 0.808 & 0.876&
			$\bigl(\begin{smallmatrix}
80&20\\ 5&95
\end{smallmatrix} \bigr)$\\
			9 & 0.964 (+1.6\%)& 0.903 & 0.672&
			$\bigl(\begin{smallmatrix}
35&65\\ 0&100
\end{smallmatrix} \bigr)$\\
			\bottomrule
		\end{tabular}
	\end{table}

	\begin{table}
		\centering
		\caption{{ROC AUC results for anomaly detection on multi-perspective MNIST~\cite{MNIST} dataset using the denoising early fusion technique (Deep SVDD w/ EF). In addition, the results of the shallow anomaly detection methods OC-SVM, KDE and IF (see Section~\ref{sec_baseline}) are listed. The numbers for Deep SVDD and IF are average values for 3 and 60 experiments, respectively.}}
		\label{tab_res_mnist_baseline}
		\setlength\extrarowheight{1pt}
		\renewcommand{\arraystretch}{1}
		\begin{tabular}{m{4cm}m{3cm}m{2cm}m{1.22cm}m{1.5cm}}
			\toprule
		{\textbf{Non-Anomalous Digits}} & {\textbf{Deep SVDD w/ EF}} & {\textbf{OC-SVM}} & {\textbf{KDE}}& {\textbf{IF}}\\
			\midrule
			
			{2} & {0.948} & {0.940} & {0.940} & {\textbf{0.952 
}}\\
			{3} & {0.941} & {\textbf{0.950}} & {0.949} & {0.942}\\
			\bottomrule
		\end{tabular}
	\end{table}

\subsection{Impact of Augmentation on Dices Dataset}

For each image in the multi-perspective dices dataset, different augmentation strategies are employed to create four different datasets (see Figure~\ref{fig_augm} in Section~\ref{sec_dices_dataset}). This section details the results obtained for different augmentation strategies. Each fusion technique is optimized using the BOHB~\cite{falkner2018bohb}. The same tuning parameters such as budget and the total number of iterations are used for all fusion techniques. The performance achieved on the respective {datasets} is summarized in Table~\ref{tab_res_augment}.


\begin{table}
		\centering
		\caption{ROC AUC scores for different augmentation techniques. A denoising autoencoder is employed in the pre-training stage of Deep SVDD~\cite{Ruff2018} for each of the fusion techniques. The results portray the performance on multi-perspective dices test dataset with augmentation. The confusion matrices are given in $\%$.}
		\label{tab_res_augment}
		\begin{tabular}{m{5cm}m{1cm}m{1.2cm}m{1cm}m{1.2cm}m{1cm}m{1.2cm}}
			\toprule
		\textbf{Dataset} & \multicolumn{2}{c}{\textbf{Early Fusion}} & \multicolumn{2}{c}{\textbf{Late Fusion}} & \multicolumn{2}{c}{\textbf{Late Fusion w/}}\\
		 & & & & & \multicolumn{2}{c}{\textbf{Dual Decoder}}\\\cmidrule{2-7}		
			& $\begin{smallmatrix}
			     \mbox{\textbf{ROC}}\\
			     \mbox{\textbf{AUC}}
\end{smallmatrix}$ & $\Big(\begin{smallmatrix}
			     \mbox{\textbf{TN}} & \mbox{\textbf{FP}} \\
			     \mbox{\textbf{FN}} & \mbox{\textbf{TP}}
\end{smallmatrix} \Big)$ & $\begin{smallmatrix}
			     \mbox{\textbf{ROC}}\\
			     \mbox{\textbf{AUC}}
\end{smallmatrix}$ & $\Big(\begin{smallmatrix}
			     \mbox{\textbf{TN}} & \mbox{\textbf{FP}} \\
			     \mbox{\textbf{FN}} & \mbox{\textbf{TP}}
\end{smallmatrix} \Big)$ & $\begin{smallmatrix}
			     \mbox{\textbf{ROC}}\\
			     \mbox{\textbf{AUC}}
\end{smallmatrix}$ & $\Big(\begin{smallmatrix}
			     \mbox{\textbf{TN}} & \mbox{\textbf{FP}} \\
			     \mbox{\textbf{FN}} & \mbox{\textbf{TP}}
\end{smallmatrix} \Big)$\\
			\midrule
			All augmentations& 0.724
&
$\bigl(\begin{smallmatrix}
84 & 16\\
53 & 47
\end{smallmatrix} \bigr)$ &0.780
&
$\bigl(\begin{smallmatrix}
30 & 70\\
10 & 90
\end{smallmatrix} \bigr)$& 0.804
&
$\bigl(\begin{smallmatrix}
88 & 12\\
33 & 67
\end{smallmatrix} \bigr)$ \\	
			w/o erased patching& 0.678
&
$\bigl(\begin{smallmatrix}
95 & 5\\
80 & 20
\end{smallmatrix} \bigr)$ & 0.774
&
$\bigl(\begin{smallmatrix}
100 & 0\\
100 & 0
\end{smallmatrix} \bigr)$ & 0.788
&
$\bigl(\begin{smallmatrix}
84 & 16\\
40 & 60
\end{smallmatrix} \bigr)$ \\
			w/o change of image constitution &0.731
&
$\bigl(\begin{smallmatrix}
99 & 1\\
87 & 13
\end{smallmatrix} \bigr)$&0.785
&
$\bigl(\begin{smallmatrix}
86 & 14\\
45 & 55
\end{smallmatrix} \bigr)$&0.796
&
$\bigl(\begin{smallmatrix}
99 & 1\\
83 & 17
\end{smallmatrix} \bigr)$\\
			w/o geometrical rot. and flipping &0.738
&
$\bigl(\begin{smallmatrix}
64 & 36\\
25 & 75
\end{smallmatrix} \bigr)$&0.762
&
$\bigl(\begin{smallmatrix}
37 & 63\\
12 & 88
\end{smallmatrix} \bigr)$&0.737
&
$\bigl(\begin{smallmatrix}
99 & 1\\
97 & 3
\end{smallmatrix} \bigr)$\\
			\bottomrule
		\end{tabular}
	\end{table}

It is important to note that the early fusion technique is not able to benefit from augmentation (no matter what flavor) but late fusion and late fusion with multiple decoders not only catches up but surpasses the performance of early fusion substantially.

\subsection{Impact of Dataset Diversity}

{As before, the goal of this work is not the classification of objects but rather the detection of anomalies on objects. We use the diversity of the Dices dataset to quantify the anomaly detection performance. The Dices dataset contains four different anomalies (cf. Figure~\ref{fig_free_fall}a). 
The starting point is the previous section where the different methods were trained with data augmentation and denoising autoencoder. Table~\ref{tab_res_diversity} shows for these cases which confusion matrices result as soon as only one of the four anomalies is considered bad and the others good.}

	\begin{table}
		\centering
		\caption{{ROC AUC scores and confusion matrices are given w.r.t. one particular type of anomaly. A denoising autoencoder and data augmentation are employed in the pre-training stage of Deep SVDD~\cite{Ruff2018} for each of the fusion techniques. The confusion matrices are given in $\%$. True positive samples (TP) are anomalous, etc.}}
		\label{tab_res_diversity}
		\renewcommand{\arraystretch}{0.1}
		\begin{tabular}{m{2cm}m{1.25cm}m{1.6cm}m{1.23cm}m{1.6cm}m{1.2cm}m{1.6cm}}
			\toprule
		{\textbf{Anomaly}} & \multicolumn{2}{c}{{\textbf{Early Fusion}}} & \multicolumn{2}{c}{{\textbf{Late Fusion}}} & \multicolumn{2}{c}{{\textbf{Late Fusion w/ Dual d.}}}\\\midrule
			& $\begin{smallmatrix}
			     \mbox{\textbf{ROC}}\\
			     \mbox{\textbf{AUC}}
\end{smallmatrix}$ & \Big($\begin{smallmatrix}
			     \mbox{\textbf{TN}} & \mbox{\textbf{FP}} \\
			     \mbox{\textbf{FN}} & \mbox{\textbf{TP}}
\end{smallmatrix}$\Big) & $\begin{smallmatrix}
			     \mbox{\textbf{ROC}}\\
			     \mbox{\textbf{AUC}}
\end{smallmatrix}$ & $\Big(\begin{smallmatrix}
			     \mbox{\textbf{TN}} & \mbox{\textbf{FP}} \\
			     \mbox{\textbf{FN}} & \mbox{\textbf{TP}}
\end{smallmatrix} \Big)$ & $\begin{smallmatrix}
			     \mbox{\textbf{ROC}}\\
			     \mbox{\textbf{AUC}}
\end{smallmatrix}$ & $\Big(\begin{smallmatrix}
			     \mbox{\textbf{TN}} & \mbox{\textbf{FP}} \\
			     \mbox{\textbf{FN}} & \mbox{\textbf{TP}}
\end{smallmatrix} \Big)$\\
			\midrule
			\includegraphics[width=2cm]{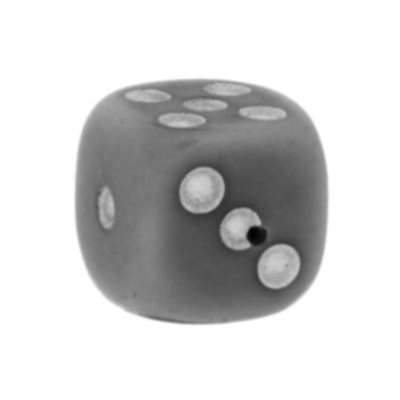}& 0.745 & $\bigl(\begin{smallmatrix}
72 & 28\\
40 & 60
\end{smallmatrix} \bigr)$ & 0.883 & 
$\bigl(\begin{smallmatrix}
23 & 77\\
0 & 100
\end{smallmatrix} \bigr)$ & 0.879 & 
$\bigl(\begin{smallmatrix}
68 & 32\\
0 & 100
\end{smallmatrix} \bigr)$ \\	\midrule
			\includegraphics[width=2cm]{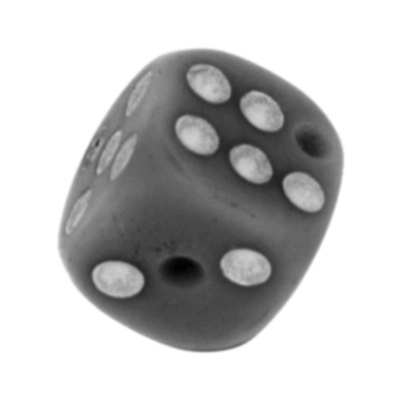}& 0.608 & 
$\bigl(\begin{smallmatrix}
73 & 27\\
55 & 45
\end{smallmatrix} \bigr)$ &  0.711 & 
$\bigl(\begin{smallmatrix}
24 & 76\\
5 & 95
\end{smallmatrix} \bigr)$ & 0.746 & 
$\bigl(\begin{smallmatrix}
71 & 29\\
20 & 80
\end{smallmatrix} \bigr)$ \\	\midrule
			\includegraphics[width=2cm]{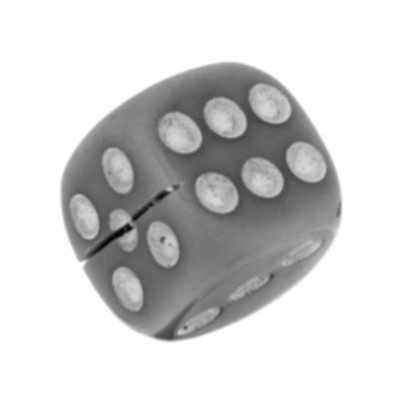}& 0.768 & 
$\bigl(\begin{smallmatrix}
72 & 28\\
40 & 60
\end{smallmatrix} \bigr)$ & 0.866 & 
$\bigl(\begin{smallmatrix}
23 & 77\\
0 & 100
\end{smallmatrix} \bigr)$ & 0.842 & 
$\bigl(\begin{smallmatrix}
67 & 33\\
10 & 90
\end{smallmatrix} \bigr)$ \\	\midrule
			\includegraphics[width=2cm]{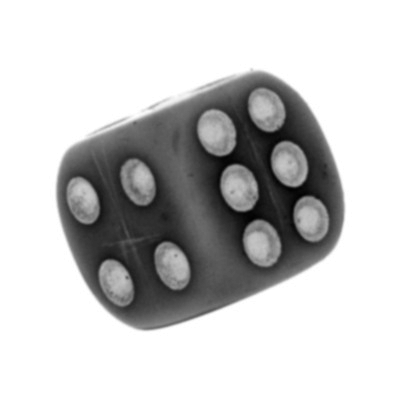}& 0.546 & 
$\bigl(\begin{smallmatrix}
71 & 29\\
65 & 35
\end{smallmatrix} \bigr)$ & 0.424 & 
$\bigl(\begin{smallmatrix}
20 & 80\\
25 & 75
\end{smallmatrix} \bigr)$ & 0.451 & 
$\bigl(\begin{smallmatrix}
61 & 39\\
75 & 25
\end{smallmatrix} \bigr)$ \\	
			\bottomrule
		\end{tabular}
	\end{table}

{The results can be roughly summarized at this point. First, the error types drilling, missing dots and sawing can be detected better than scratches. Second, high TP rates are achieved with the late fusion techniques compared to the early fusion.}
{Hence, With the numbers in Table \ref{tab_res_diversity} we show a dependency of our approaches to the available types of anomalies of the dices dataset as the anomaly detection rate is dependent what anomaly is to be detected. }

\section{Discussion}
\label{sec_dis}

While analysing the results for the multi-perspective anomaly detection techniques (Section~\ref{sec_res_multi_persp}) specifically the ones without any augmentation (Tables~\ref{tab_res_multi_persp} and~\ref{tab_res_multi_persp_denoise}), it is clear that the early fusion technique performs the best. The reason for this is based on the way that the data points are mapped in the feature space where the hypersphere is found. As described in Section~\ref{sec_fusion}, in early fusion there is no averaging of the feature space vectors compared to the other two fusion techniques. The different perspectives are stacked together and fed as input to the algorithm. In late fusion and late fusion with dual decoders, due to averaging of the feature vectors, the overall location is transferred to the wrong side of the hypersphere. For example, consider a case wherein one perspective the anomaly is present and in the opposite perspective, it is not. Hence, the feature space vector for the anomalous perspective is mapped outside the hypersphere and for the non-anomalous perspective inside. In particular cases, the averaging process puts the final feature vector inside the hypersphere which would indeed lead to misclassification. {As soon as more than two perspectives are used for the evaluation, this effect should worsen.}

From Table~\ref{tab_res_multi_persp_denoise}, it can be observed that the late fusion technique performs better than the late fusion with the dual decoder. This can be attributed to the same averaging phenomenon: In numerical values, the average Euclidean distance between feature vectors $\phi_1$ and $\phi_2$ for late fusion with and without dual decoder strongly deviate from each other by the factor of 100.
As mentioned before, the hyperparameters are optimized w.r.t. ROC AUC scores and not, e.g., average precision. Hence, the confusion matrices do not look satisfactorily in particular cases, e.g., late fusion w/ dual decoder.

We evaluated the best performing fusion technique, i.e., early fusion with denoising autoencoder on the extended multi-perspective MNIST dataset. The results achieved are summarized together in Table~\ref{tab_res_mnist} and good performance is observed for all the different datasets. The dataset having digit one as a non-anomalous class achieved the highest ROC AUC score of more than 99\%. Compared to ``single-perspective'' anomaly detection in \cite{Ruff2018}, ROC AUC scores are further increased relatively up to 11\% for MNIST digit five.  

From Figure~\ref{fig_mnist_inference}a it can be analyzed that the denoising early fusion technique predicts the non-anomalous samples well as digit zero is not present in the top-five anomalous samples. It is also interesting to analyze the fifth most non-anomalous sample which in reality is an example of a misclassified sample. This sample has an image of digit five in channel one and digit six in channel two. One reason for this misclassification could be because of the high visual similarity of the image representing digit six with the non-anomalous class samples containing zeros. In contrast to that, the correct classified image stack having digit six in channel one appears far dissimilar to zero with a closed circle. 
In Figure~\ref{fig_mnist_inference}c it is visible that the autoencoder used in the pre-training stage of the denoising early fusion technique fails to reconstruct the anomalous samples.

The high performance of our fusion techniques on the MNIST data compared to the performance on our dices dataset (ROC AUC scores of up to 99\% vs. 74\%) can be attributed to two arguments. First, the anomaly itself covers a greater area in the images (the whole digit is different vs. small scratch on dice) which leads to a higher impact on the used error function during training. In the literature, this aspect is known as ``high-level'' and ``low-level'' anomalies~\cite{Ruff2021}. {In the experiment with MNIST, it should be emphasized that the network used is capable of learning the different digits as the respective good object.} In the dices dataset, an anomaly is usually present in only one of the perspectives. Whereas, in the MNIST multi-perspective dataset, often both perspectives hold different \mbox{``anomalous'' digits.}

{It is precisely at this point that it is important to draw comparisons with state-of-the-art shallow anomaly detection algorithms. Looking at the results in Table \ref{tab_res_mnist_baseline}, in the case of the ``high-level'' anomalies of the multi-perspective MNIST dataset it makes sense to use an Isolation Forest or One-Class SVM algorithm for the evaluation. However, as soon as ``low-level'' anomalies occur, the Deep SVDD w/ fusion approach should be chosen (see Table \ref{tab_res_multi_baseline}). Obviously, extracting features with a deep network is likely to perform better than with PCA.}
{Furthermore, the performance of the methods depends on the diversity of the dataset, i.e., which ``low-level'' type of anomaly is to be detected.} {While adding another dataset would benchmark diversity and thus generalization more, important properties can already be inferred.} {A closer look at Table \ref{tab_res_diversity} shows that types drilling and sawing are detected much better than, e.g., scratches, since correspondingly less area in the image is occupied by the anomaly. Interestingly, the detection of missing dots works less well than drilling. One reason may be that the network does not learn complex dot patterns of the individual cube faces and is therefore not sensitive to missing dots in the \mbox{individual pattern}.}

Observing the results from the ablation study performed with the different augmented multi-perspective dices datasets, it is clear that the augmentations help the model to perform better compared to previous ROC AUC scores. For example in the first set, where all the different types of augmentations are employed, an ROC AUC of $80\%$ (see \mbox{Table~\ref{tab_res_augment}}) is achieved which is the best so far for multi-perspective anomaly detection on the dices test dataset. One may notice that the augmentation technique of erasing patches has similarities with the appearance of certain anomalies in the testing dataset, i.e., drilled holes and missing dots. As mentioned in \cite{Goodfellow-et-al-2016}, this may lead to misclassification. In our case, the erased patches are too big and have a rectangular shape compared to particular anomalies. Hence, the application of this augmentation technique helps the model to generalize better.

It is important to note that the early fusion is not able to benefit from the augmentations (no matter what flavor), but the late fusion and late fusion with multiple decoders not only catches up but surpasses the performance of the early fusion approach. {Compare also \mbox{Table \ref{tab_res_diversity}} in this context.} {Thus, we have confirmed an important result of Seeland and \mbox{Mäder \cite{Seeland2021}} also for an unsupervised setting, that fusion should occur as late as possible in latent space.} Data augmentation has regularization effects and a high extent of regularization may also cause the model to underfit, which could be the case for early fusion. Interestingly, the misclassification issue while averaging two feature vectors is solved here which leads to the conclusion that as soon as the extraction of relevant features works after training with augmented and noised data, the averaging issue is of less relevance. The best performance of the dual decoding network of $80\%$ ROC AUC score can be attributed to the same aspect. The feature extraction of the encoder should work effectively so that the two decoders can process the information successfully. 
Furthermore, it is important to note that all the techniques that boost the multi-perspective anomaly detection performance can also be employed for single-perspective anomaly detection based on Deep SVDD.


\section{Conclusions}
\label{sec_con}

This work presents novel approaches that address multi-perspective anomaly detection using the fusion of information to extend the well-known Deep SVDD algorithm. As a proof of concept, we employ these approaches for two perspectives. {In the case the two perspectives have a chance to miss the anomaly at all it is recommended to use more than two perspectives.} The core idea is to fuse the information of the two perspectives at the input of the network (``early fusion'') and the output (``late fusion''). The network itself---a convolutional autoencoder with or without noised input---is trained from scratch whereas hyperparameters were optimized with the Bayesian toolset BOHB.

Another contribution of this work is the novel multi-perspective dices dataset which we use for evaluating the proposed fusion techniques. One promising fusion method is employed on the standard MNIST dataset. This dataset is successfully transferred into the multi-perspective setting and evaluated using early fusion with the denoising autoencoder technique. Overall, the results on the adapted MNIST dataset illustrate the robustness and effectiveness of our approach {on ``high-level'' anomalies, independent what object (digit) is considered}. A ROC AUC score of >99\% for one of the 10 MNIST adaptions is achieved. The highest ROC AUC score (80\%) while predicting anomalies in the multi-perspective dices dataset is achieved with the late fusion technique together with different image augmentation strategies. {Here, the detailed anomaly detection rate strongly depends on what kind of anomaly is to be detected (e.g., scratches vs. missing dots on dices).}  

For future work, increasing the size of the training set size can improve the performance as the authors in \cite{Goodfellow-et-al-2016} show that the classical supervised classification networks require approximately 5000 labeled examples per class to match or exceed human performance. Augmenting data is inexpensive compared to human effort and time; the number and types of augmentations can be increased to further improve the performance of multi-perspective anomaly detection. Furthermore, in the use case described in this work, there is a lack of anomalous samples available during the training process and another interesting approach would be to consider the augmented samples as anomalous for forming an adversarial training process. The augmented samples can be used indirectly in the training process by not updating the model on these augmented samples but using them as anomalous samples. Utilizing a similarity metric, the difference between the good class samples and the augmented samples can be maximized which will indeed force the network to extract even more robust features from the good class samples.\nocite{kohanbash2011wireless}

%
\vspace{6pt} 




\funding{This work was supported by the Fraunhofer Internal Programmes under Grant No. WISA 833 959.}

\acknowledgments{We sincerely thank the Inline Vision Systems group for the discussions and the provision of the experimental platform. We thank Oier Mees for the discussions.}


\abbreviations{The following abbreviations are used in this manuscript:\\

\noindent 
\begin{tabular}{@{}ll}
AE & Autoencoder\\
BOHB & Bayesian Optimization and Hyperband \cite{falkner2018bohb}\\
CAE & Convolutional Autoencoder\\
DAE & Denoising Autoencoder\\
Deep SVDD & Deep Support Vector Data Description\\
DOAJ & Directory of open access journals\\
FC & Fully Connected\\
GIST & "The 'gist' is an abstract representation of the scene that spontaneously activates memory representations of scene categories (a city, a mountain, etc.)" \cite{Oliva2001}\\
GPU & Graphical Processing Unit\\
HOG & Histogram of Oriented gradient\\
IPM & Institute for Physical Measurement Techniques, Freiburg, Germany\\
KDE & Kernel Density Estimation\\
LBP & Local Binary Patterns \\
MDPI & Multidisciplinary Digital Publishing Institute\\
MNIST & Modified National Institute of Standards and Technology\\
OC-SVM & One Class Support Vector Machine\\
ROC & Receiver Operating Characteristic\\
ROC AUC & Area Under the Receiver Operating Characteristic Curve\\
SVM & Support Vector Machine \\
TN, TP, FN, FP & True Negative, etc.
\end{tabular}}


\appendixtitles{no} 
\appendix
\section{}

\subsection{Evaluation Metrics}
\label{sec_calc_metrics}

For the dices dataset, true labels (0 for the non-anomalous class and 1 for the anomalous class) are present for all the samples in the test dataset. The anomaly scores are computed by subtracting network output with the center and then with the radius (see Equation~\ref{eq_anomaly_score}) in order evaluate if the novel point lies inside the hypersphere. These scores can be directly used to compute the ROC AUC score using the scikit-learn library~\cite{scikit-learn}. However, for computing the other metrics i.e. precision and recall, the anomaly scores have to be converted into labels (0 or 1). This is performed considering the sign of the anomaly score i.e. for a negative score, the predicted label is zero and otherwise one. The precision and recall is computed using the macro-average functionality available in the scikit-learn library~\cite{scikit-learn}. Macro-averaging implies that each class is considered independently. The metrics are first computed for each class and then averaged over the total classes. This helps in getting rid of any effects of data imbalance. The precision, recall, and confusion matrix values are chosen from the best performing experiment. The values in the confusion matrix are normalized and converted into percentages. True positive (TP) correspond to true anomalous samples etc.

\subsection{Hyperparameters}
\label{sec_app_hyper_multi}

\begin{table}[H]
	\centering
	\caption{List of hyperparameters used for training the fusion techniques for multi-perspective anomaly detection on our multi-perspective dices dataset (without augmentations). AE refers to the autoencoder from the pre-training stage.}
	\label{tab_app_hyperparameter_multi_view}
	\setlength\arrayrulewidth{0.6pt}
	\renewcommand{\arraystretch}{1.5}
	\begin{tabular}{m{3.5cm}| m{3cm}  m{3cm} m{3cm}}
		\toprule
		\textbf{Hyperparameters} & \textbf{Early fusion} & \textbf{Late fusion} & \textbf{Late fusion with dual decoders}\\
		\midrule
		Optimizer AE & Adam & Adam & Adam \\
		Batch size AE& 23 & 21&24 \\ 
		Learning rate AE &0.00657 & 0.000178 & 0.000627\\
		Weight decay AE& 3.346e-08 & 4.8869e-07 & 4.88e-08 \\
		Epochs AE & 127 & 28 & 127 \\
		Batch size & 11 & 28 & 11\\
		Learning rate & 7.706e-06 & 2.339e-05 & 7.706e-05\\
		Weight decay & 1.3025e-09 & 4.95e-08 & 1.3025e-09\\
		Epochs & 80 & 84 & 80\\
		Feature space & 640 & 392 &392\\
		$\nu$& 0.4 & 0.4 & 0.4\\
		Optimizer & Adam & Adam & Adam \\
        \bottomrule
	\end{tabular}
\end{table}


\reftitle{References}


\externalbibliography{yes}
\bibliography{template}





\end{document}